\documentclass[11pt]{article}

\usepackage[preprint]{acl}

\usepackage{times}
\usepackage{latexsym}

\usepackage[T1]{fontenc}
\usepackage[utf8]{inputenc}

\usepackage{microtype}

\usepackage{inconsolata}

\usepackage{graphicx}

\usepackage{amsmath}
\usepackage{amssymb}
\usepackage{mathtools}
\usepackage{amsthm}
\usepackage{microtype}
\usepackage{graphicx}
\usepackage{subcaption}
\usepackage{booktabs}
\usepackage[ruled,vlined]{algorithm2e}
\usepackage{xcolor}
\usepackage[most]{tcolorbox}
\definecolor{myred}{RGB}{255, 230, 230}
\definecolor{mygreen}{RGB}{230, 255, 230}
\definecolor{myblue}{RGB}{220, 230, 255}
\definecolor{darkred}{RGB}{180, 0, 0}
\definecolor{darkgreen}{RGB}{0, 100, 0}
\definecolor{darkblue}{RGB}{0, 0, 150}

\title{Deep Dense Exploration for LLM Reinforcement Learning via Pivot-Driven Resampling}

\author{
  Yiran Guo\textsuperscript{1} \quad
  Zhongjian Qiao\textsuperscript{2} \quad
  Yingqi Xie\textsuperscript{2} \quad
  Jie Liu\textsuperscript{1} \quad
  Dan Ye\textsuperscript{1} \\
  \textbf{Ruiqing Zhang}\textsuperscript{3} \quad
  \textbf{Shuang Qiu}\textsuperscript{2}\thanks{\ \ Corresponding authors.} \quad
  \textbf{Lijie Xu}\textsuperscript{1}\footnotemark[1] \\
  \\
  \textsuperscript{1}Institute of Software, Chinese Academy of Sciences \\
  \textsuperscript{2}City University of Hong Kong \quad \textsuperscript{3}Baidu \\
  \texttt{shuanqiu@cityu.edu.hk}, \texttt{xulijie@iscas.ac.cn}
}

\begin{document}
\maketitle
\begin{abstract}
    Effective exploration is a key challenge in reinforcement learning for large language models: discovering high-quality trajectories within a limited sampling budget from the vast natural language sequence space. Existing methods face notable limitations: GRPO samples exclusively from the root, saturating high-probability trajectories while leaving deep and error-prone states under-explored. Tree-based methods disperse budgets across trivial or unrecoverable states, causing sampling dilution that fails to uncover rare correct suffixes and destabilizes local baselines. To address this, we propose Deep Dense Exploration (DDE), a strategy that focuses exploration on \textit{pivots}—deep, recoverable states within unsuccessful trajectories. We instantiate DDE with DEEP-GRPO, which introduces three key innovations: (1) a utility-guided pivot sampling distribution that balances depth-based branching value with online recoverability estimation; (2) local dense resampling at each pivot to increase the probability of discovering correct subsequent trajectories; and (3) a dual-stream optimization objective that decouples global policy learning from local corrective updates. Experiments on mathematical reasoning benchmarks and multi-hop QA agent tasks demonstrate that our method consistently outperforms GRPO, tree-based methods, and other strong baselines. Code is available at \url{https://github.com/AgentCombo/DEEP-GRPO}.
\end{abstract}

\section{Introduction}

\begin{figure*}[t]
    \centering
    \includegraphics[width=\textwidth]{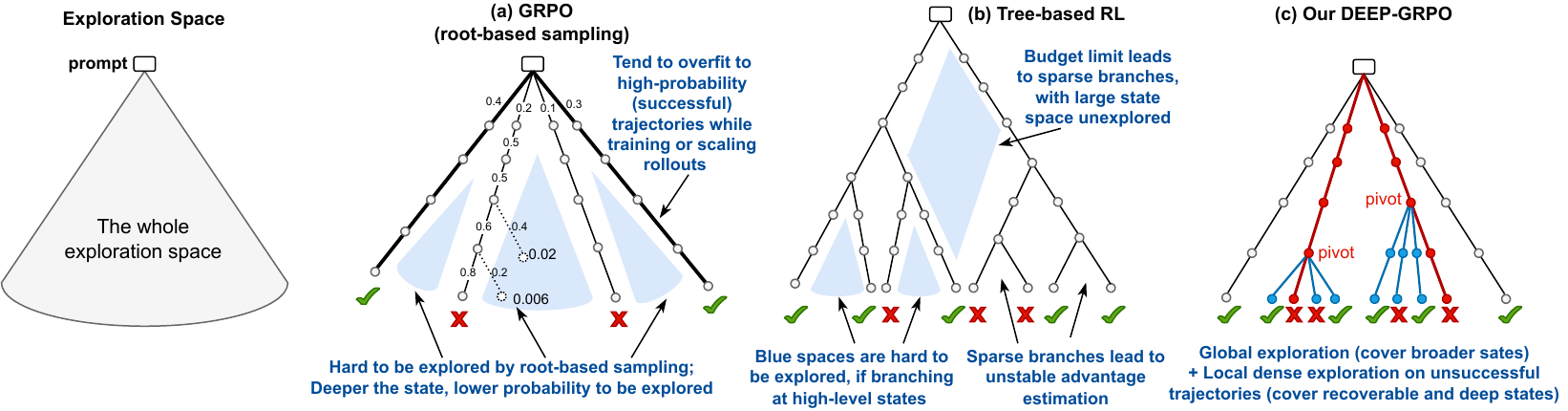}
    % {figures/compare_grpo_treerl.pdf}
    \caption{Comparison of Exploration Strategies. (a) GRPO scales up rollouts from the root, wasting budget on redundant high-probability paths while failing to explore deep states. (b) Tree-based methods perform dispersed branching with sparse samples, leading to limited local exploration and unstable local advantage estimation. (c) Our DEEP-GRPO identifies critical ``pivot" states and concentrates dense resampling there. We decouple optimization into global and local streams (with gradient masking), combining global policy learning with local refinement.}
    \label{fig:compare_grpo_treerl}
\end{figure*}

Reinforcement Learning (RL) has become a key method for enhancing the reasoning capabilities of Large Language Models (LLMs)~\cite{deepseekai2025deepseekr1incentivizingreasoningcapability, kimiteam2025kimik15scalingreinforcement, qwen3technicalreport}. A critical challenge within this paradigm is achieving effective exploration in the vast and complex search space of natural language reasoning. This challenge is compounded by the strict computational budgets typical of online RL training, where often only a small number of rollouts (e.g., 8–16) are feasible per prompt. Under this constraint, a fundamental resource allocation problem emerges: how can we strategically distribute these limited sampling resources to explore the most informative states for continuous improvement in model capabilities?

\begin{figure}[t]
    \centering
    \includegraphics[width=0.9\linewidth]{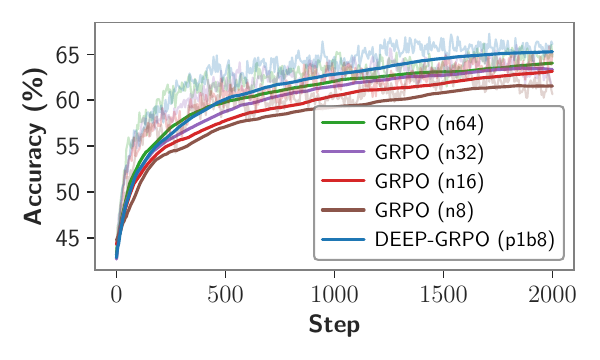}
    \caption{Diminishing Returns of Naive Scaling. We evaluate GRPO on GSM8K by scaling the number of rollouts per prompt ($N$) from 8 to 64. While performance improves from $N=8$ to $N=16$, it quickly saturates with negligible gains for $N=32$ and $N=64$. In contrast, DEEP-GRPO ($P_1B_8$) selects a pivot from a failed trajectory and branches 8 new rollouts from that point, effectively converting the additional exploration budget into a larger performance improvement.}
    \label{fig:diminishing_returns}
\end{figure}

Current mainstream approaches, particularly Group Relative Policy Optimization (GRPO)~\cite{shao2024deepseekmathpushinglimitsmathematical}, rely on sampling complete trajectories from the root to drive exploration. However, this paradigm faces significant limitations. First, it suffers from exploration sparsity. This policy naturally favors high-probability tokens, making potentially valuable states on lower-probability trajectories statistically difficult to access, especially those deep states. Second, as training progresses, the policy rapidly overfits to already mastered successful trajectories, leading to a sharp drop in exploration entropy~\cite{yu2025dapoopensourcellmreinforcement,cui2025entropy}. This premature convergence constrains the model's capacity to discover novel solutions. Consequently, simply increasing the number of root-level rollouts yields diminishing returns (as shown in Figure \ref{fig:diminishing_returns}), as the computational budget is inefficiently wasted on redundant, high-confidence trajectories rather than penetrating deep, unexplored states. 
% Furthermore, GRPO's coarse-grained, trajectory-level advantage estimation erroneously penalizes valid reasoning steps embedded within failed trajectories~\cite{ma2023letsrewardstepstep}, further hindering the learning process.

To enhance exploration depth, recent studies have introduced tree-based RL methods~\cite{hou2025treerl, liu2025attention, ji2025tree, zheng2025first} that initiate branching exploration from intermediate trajectory states. However, under the strict computational budget constraints of online RL, these approaches encounter an intrinsic sample dispersion challenge. By dispersing the limited sampling budget across various intermediate states identified by heuristic metrics, they induce extreme sample sparsity at individual branch points. This sparsity results in highly limited local exploration and renders the computation of stable local advantage estimates infeasible (see Figure~\ref{fig:compare_grpo_treerl} and the motivating example in Appendix~\ref{app:motivating_example}). Consequently, to update the policy, these methods aggregate heterogeneous trajectories from disparate branches. This practice introduces significant bias by conflating the model's natural output distribution with artificially induced exploration paths, ultimately undermining training stability—a phenomenon we empirically verify in Appendix~\ref{app:tree_instability}.

To overcome the aforementioned limitations, we introduce Deep Dense Exploration (DDE), a strategy that complements broad root-level sampling with targeted deep exploration. The core idea is to concentrate auxiliary rollouts on ``pivots''—critical states embedded within failed trajectories that are deep-seated yet recoverable. Our approach is grounded in two key insights: (1) \textbf{\textit{Recoverability of Failed Trajectories}}: Many failed trajectories contain valid reasoning prefixes and remain recoverable. By prioritizing resampling from such states, we can discover correct completions that form high-quality contrastive pairs with the original errors. (2) \textbf{\textit{Complementarity to Root Sampling}}: Deep states are exponentially harder to access via root sampling. Targeting these deep regions provides learning signals that are complementary to root sampling. Moreover, as highlighted by~\citet{deng2025trial}, tokens in these later stages often exhibit higher uncertainty, thus requiring intensive optimization.

Based on these insights, we propose DEEP-GRPO, a novel RL algorithm that implements the Deep Dense Exploration (DDE) strategy. Our method preserves global breadth through root-level sampling while introducing targeted local exploration at selected pivots via three integrated components. First, we employ a 
\textbf{\textit{Utility-guided Pivot Sampling Distribution}} to select pivots from failed trajectories by balancing depth-based branching value with online recoverability estimation. From these pivots, we execute \textbf{\textit{Dense Local Resampling}}, concentrating auxiliary rollouts to increase the chance of discovering correct suffixes and to compute stable local advantages. Finally, we introduce \textbf{\textit{Dual-Stream Optimization}}, which combines the standard GRPO objective on root-sampled rollouts with a local objective derived from auxiliary chains while masking gradients on shared prefixes. Experiments on mathematical reasoning and multi-hop QA agent benchmarks show that DEEP-GRPO consistently outperforms GRPO, tree-based exploration methods, and other strong baselines.

Our contributions are summarized as follows:

\begin{itemize}
  \item We propose Deep Dense Exploration (DDE), which preserves exploration breadth through root-level sampling while adding targeted deep and dense exploration at pivots within failed trajectories.

  \item We instantiate DDE as DEEP-GRPO, combining a utility-guided pivot sampling distribution, dense local resampling, and dual-stream optimization with prefix gradient masking.

  \item We validate DEEP-GRPO on mathematical reasoning and multi-hop QA agent benchmarks, showing consistent gains over GRPO, tree-based methods, and other strong baselines.
\end{itemize}

\section{Related Work} 

In this section, we focus on the most closely related work on structured exploration. Additional related work is provided in Appendix~\ref{app:more_related_work}. Inference-time methods, such as RAP~\cite{hao2023reasoning} and LATS~\cite{zhou2023language}, perform extensive tree search (e.g., MCTS~\cite{silver2017mastering}) to uncover solutions during generation. However, their prohibitively high computational cost renders them impractical for the inner loop of online RL training. Offline training methods, such as ReST-MCTS~\cite{zhang2024restmctsllmselftrainingprocess} and AlphaMath~\cite{chen2024alphamathzeroprocesssupervision}, utilize tree search to synthesize high-quality trajectories for iterative Supervised Fine-Tuning (SFT). While effective, these approaches rely on decoupled data generation cycles rather than continuous online optimization. Online RL methods, therefore, attempt to integrate structure directly into the policy update loop. Existing approaches, including TreeRL~\cite{hou2025treerl}, FR3E~\cite{zheng2025first}, and AttnRL~\cite{liu2025attention}, branch from intermediate steps selected by heuristics. Specifically, TreeRL and FR3E target high-entropy tokens, while AttnRL focuses on high attention scores, and Tree-GRPO~\cite{ji2025tree} employs random branching. However, these heuristics prioritize different attributes than learning potential: high entropy often stems from trivial linguistic synonyms rather than logical uncertainty, while high attention signals step importance regardless of correctness. In contrast, our method targets error-prone states—specifically those deep within trajectories that are hard to reach via root sampling—to provide complementary learning signals. Furthermore, in terms of optimization, prior methods typically couple root-sampled and branched trajectories within the same loss. Such coupling may introduce instability, since their disparity in sample quantities can lead to imbalanced weighting in the loss. DEEP-GRPO addresses this by decoupling the optimization of main and auxiliary chains, preventing such interference.

\section{Preliminary}
\label{sec:preliminary}

We consider outcome-supervised RL for language generation. Given a query $\mathbf{x}$, the policy $\pi_\theta$ generates a trajectory $\tau=(w_1,\ldots,w_T)$ and receives a verifiable terminal reward $R(\tau,\mathbf{x})$, e.g., $1$ for a correct answer and $0$ otherwise. 
DEEP-GRPO builds on GRPO~\cite{shao2024deepseekmathpushinglimitsmathematical}, which samples a group of $G$ trajectories from $\pi_{\theta_{\text{old}}}$ and optimizes a clipped policy-gradient objective without a value model. 
For each trajectory $\tau^i$, GRPO computes a group-relative advantage
$A^i=(R(\tau^i)-\mu_{\text{group}})/\sigma_{\text{group}}$, where $\mu_{\text{group}}$ and $\sigma_{\text{group}}$ are the mean and standard deviation of rewards within the group. 
We use $\rho_t^i=\pi_\theta(w_t^i|\mathbf{x},w_{<t}^i)/\pi_{\theta_{\text{old}}}(w_t^i|\mathbf{x},w_{<t}^i)$ to denote the token-level probability ratio.

\section{DEEP-GRPO}

\begin{figure}[ht]
    \centering
    \includegraphics[width=0.95\linewidth]{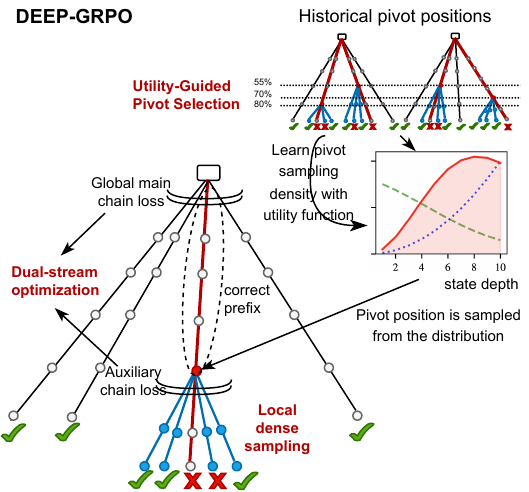}
    \caption{Method Overview.}
    \label{fig:hierarchical_trajectory_generation}
\end{figure}

In this section, we present DEEP-GRPO, a novel method designed to enhance the exploration of LLM reinforcement learning. Unlike GRPO which explores exclusively from the root, or tree-based methods that disperse resources across intermediate states, our approach concentrates the exploration budget on critical pivots—states that are difficult to reach yet offer high-value learning signals. We hypothesize that targeted exploration at these critical states significantly enhances model robustness and drives continuous improvement in the later stages of training. Figure \ref{fig:hierarchical_trajectory_generation} provides a holistic view of DEEP-GRPO. We first describe how trajectories are segmented into candidate branching points (Sec. \ref{subsec:trajectory_segmentation}). Next, we introduce our utility-guided pivot selection (Sec. \ref{subsec:pivot_selection}), followed by the hierarchical generation of main chains and auxiliary chains (Sec. \ref{subsec:hierarchical_trajectory_generation}). Finally, we derive a dual-stream optimization objective that integrates local refinement objectives into the global policy learning (Sec. \ref{subsec:optimization_objective}). The complete training procedure is summarized in Algorithm \ref{alg:tb_grpo}.

\subsection{Trajectory Segmentation}
\label{subsec:trajectory_segmentation}

To create a discrete and manageable set of potential exploration points, we first segment the continuous token stream of a trajectory into a sequence of $T$ candidate branching points. This reduces the search space for our pivot selection mechanism compared to token-level branching. For reasoning tasks, we segment trajectories either semantically (e.g., by sentence delimiters like `.') or via fixed-length token chunking. The latter allows for flexible control over segment granularity, enabling fine-grained intervention regardless of the semantic structure. For agent scenarios, we segment trajectories according to the natural interaction structure, treating each \textit{Think--Action--Observation} step as a candidate branching point.

\subsection{Utility-Guided Pivot Sampling}
\label{subsec:pivot_selection}

Unlike prior tree-based methods, DEEP-GRPO performs auxiliary branching only on unsuccessful trajectories, because branching from failures provides a critical opportunity to recover correct suffixes from previously incorrect paths. This generates high-value correction signals, enabling the model to learn effectively by contrasting the original failure with the newly found success. In contrast, successful trajectories already provide positive learning signals, and further branching from them often offers limited additional benefit.

For branching point selection, existing tree-based RL methods choose positions at random~\cite{ji2025tree} or using heuristic criteria such as entropy~\cite{hou2025treerl} or attention~\cite{liu2025attention}. In contrast, we construct a pivot sampling distribution by considering two factors: the likelihood of recovering a correct suffix from a prefix, and the learning value of branching from that position if recovery succeeds. The first factor avoids allocating rollouts to prefixes with low expected recovery potential, while the second favors positions whose continuations can provide useful learning signals beyond standard root rollouts. We combine these two factors in a utility-guided pivot sampling distribution:
\begin{equation}
    Q(t) \propto W(t) \cdot R(s_{<t}),
\label{eq:proposal_dist}
\end{equation}
where $s_{<t}$ denotes the prefix ending at the $t$-th candidate branching point. Here, $W(t)$ captures the branching value of selecting position $t$, while $R(s_{<t})$ captures the recoverability of the prefix. The multiplicative form acts as a soft conjunction, assigning high probability only to positions that are both valuable to branch from and likely to be recoverable. This formulation also provides a unified view of several exploration behaviors: different choices of $W(t)$ and $R(s_{<t})$ can mimic root sampling, random branching, or depth-biased branching. We visualize these representative cases in Appendix~\ref{app:pivot_sampling_configs}.

In this work, we adopt simple yet effective instantiations for both components. For the branching-value term, we use relative depth $r_t=t/T$ as a lightweight proxy, where $T$ is the total number of candidate branching points defined in Sec.~\ref{subsec:trajectory_segmentation}. 
Deeper prefixes are less likely to be reached by root sampling, and branching from them can therefore provide learning signals that better complement those obtained from root rollouts. We instantiate the branching-value term with a polynomial depth bias:
\begin{equation}
    W(t)=r_t^\gamma,
\end{equation}
where $\gamma \ge 0$ controls the preference for deeper pivots.

For the recoverability term, we also use relative depth as a lightweight proxy for recovery potential. The intuition is that, in an unsuccessful trajectory, a later prefix is more likely to contain an irreversible error, making it harder for auxiliary rollouts to recover a correct suffix. We therefore instantiate this term with a lightweight logistic estimator:
\begin{equation}
    R_\phi(s_{<t})=\sigma(w \cdot r_t + b).
\end{equation}
The estimator $R_\phi$ is updated online from historical branching outcomes. Details on the online update of $R_\phi$ are provided in Appendix~\ref{app:rphi_update}, and statistical evidence for the depth--recoverability relationship is provided in Appendix~\ref{app:depth_recoverability}.

\subsection{Hierarchical Trajectory Generation}
\label{subsec:hierarchical_trajectory_generation}

Leveraging the trajectory segmentation defined in Sec.~\ref{subsec:trajectory_segmentation} and the pivot selection strategy in Sec.~\ref{subsec:pivot_selection}, we implement a two-stage hierarchical generation process:

\textbf{Main Chain Sampling ($\mathcal{T}_{\text{main}}$).}
We first sample a group of $G$ trajectories from the root following standard GRPO: $\mathcal{T}_{\text{main}} = \{\tau^1, \dots, \tau^G\}$, where each $\tau^i \sim \pi_\theta(\cdot|\mathbf{x})$. We identify the subset of unsuccessful trajectories as $\mathcal{T}_{\text{fail}} \subset \mathcal{T}_{\text{main}}$.

\textbf{Auxiliary Chain Sampling ($\mathcal{T}_{\text{aux}}$).}
For each failed trajectory $\tau^i \in \mathcal{T}_{\text{fail}}$, we segment it into $T$ candidate branching points following the strategy in Sec.~\ref{subsec:trajectory_segmentation}. We then sample a specific pivot step $t^*_i \in \{1, \dots, T\}$ according to the distribution $\mathcal{Q}(t)$ defined in Eq.~\ref{eq:proposal_dist}.

Given the prefix $s_{<t^*_i}$ (the partial trajectory of $\tau^i$ ending at the chosen branching point), we perform dense local resampling to generate $K$ auxiliary completions, forming the set $\mathcal{T}_{\text{aux}}^{(i)} = \{\hat{\tau}^{i,1}, \dots, \hat{\tau}^{i,K}\}$:
\begin{equation*}
    \hat{\tau}^{i,k} = s_{<t^*_i} \circ \mathbf{y}^{i,k}, \quad \text{where } \mathbf{y}^{i,k} \sim \pi_\theta(\cdot | s_{<t^*_i}, \mathbf{x})
\end{equation*}
Here, $\circ$ denotes concatenation, and $\mathbf{y}^{i,k}$ represents the newly generated suffix tokens. As shown in Figure \ref{fig:hierarchical_trajectory_generation}, this process constructs a ``bifurcation'' in the reasoning path, forcing the model to explore alternative solutions specifically from error-prone pivot states rather than restarting from the root.

We provide a comparison of computational efficiency between our method and tree-based approaches in Appendix~\ref{app:computational_efficiency}.

\subsection{Dual-Stream Optimization Objective}
\label{subsec:optimization_objective}

Our hierarchical generation process yields two distinct sets of trajectories: main chains $\mathcal{T}_{\text{main}}$ sampled from the root ($\mathbf{x} \sim \mathcal{D}$), and auxiliary chains $\mathcal{T}_{\text{aux}}$ sampled from specific pivot states. How should we integrate both data sources to effectively optimize the policy?

A naive approach would be to merge both sets into the same optimization batch. However, this introduces significant sample imbalance and weight instability. The volume of training samples for both streams fluctuates dynamically throughout training. Putting them together would cause the gradient contribution of each stream to shift unpredictably; if one stream produces a disproportionately large number of samples, it would drown out the optimization signal from the other stream, destabilizing the training process.

To address this, we propose a Dual-Stream Optimization strategy. We decouple the two streams, treating the optimization of auxiliary chains as a supplementary task to aid policy learning. This separation allows us to explicitly control the influence of the auxiliary task via a hyperparameter $\lambda$. The total objective $\mathcal{J}(\theta)$ minimizes the weighted hybrid loss:
\begin{equation}
\begin{aligned}
    \mathcal{J}(\theta) &= \mathbb{E}_{\mathbf{x} \sim \mathcal{D}} \Bigg[ \frac{1}{G} \sum_{i=1}^G \mathcal{L}_{\text{main}}(\tau^i) \\
    &\quad + \lambda \frac{1}{|\mathcal{T}_{\text{fail}}|} \sum_{\tau^i \in \mathcal{T}_{\text{fail}}} \frac{1}{K} \sum_{k=1}^{K} \mathcal{L}_{\text{aux}}(\hat{\tau}^{i,k}) \Bigg]
\end{aligned}
\label{eq:optimization_objective}
\end{equation}
where $\lambda$ balances the primary global policy learning with the auxiliary local error correction.

\textbf{Main Chain Loss ($\mathcal{L}_{\text{main}}$).}
For the main stream, for each trajectory $\tau^i \in \mathcal{T}_{\text{main}}$, we compute the global advantage $A^i_{\text{global}}$ using the group-wise statistics of $\mathcal{T}_{\text{main}}$. The loss is:
\begin{equation*}
\begin{aligned}
    \mathcal{L}_{\text{main}}(\tau^i) &= \frac{1}{|\tau^i|} \sum_{t=1}^{|\tau^i|} \biggl[ \min \Bigl( \rho_t^i A^i_{\text{global}}, \\
    & \text{clip}(\rho_t^i, 1-\epsilon, 1+\epsilon) A^i_{\text{global}} \Bigr) - \beta \mathbb{D}_{\text{KL}} \biggr]
\end{aligned}
\end{equation*}
\textbf{Auxiliary Chain Loss ($\mathcal{L}_{\text{aux}}$).}
For the auxiliary stream, we compute the local advantage $A^{i,k}_{\text{local}}$ for each branch $\hat{\tau}^{i,k}$ relative only to its sibling branches in $\mathcal{T}_{\text{aux}}^{(i)}$. Crucially, we apply gradient masking to freeze the prefix $s_{<t^*_i}$, calculating the loss solely on the generated suffix $\mathbf{y}^{i,k}$:
\begin{equation*}
\begin{aligned}
    \mathcal{L}_{\text{aux}}(\hat{\tau}^{i,k}) &= \frac{1}{|\mathbf{y}^{i,k}|} \sum_{w_t \in \mathbf{y}^{i,k}} \Bigg[ \min \Big( \hat{\rho}_t^{i,k} A^{i,k}_{\text{local}}, \\
    &\hspace{-0.6em} \text{clip}(\hat{\rho}_t^{i,k}, 1-\epsilon, 1+\epsilon) A^{i,k}_{\text{local}} \Big) - \beta \mathbb{D}_{\text{KL}} \Bigg]
\end{aligned}
\end{equation*}

\section{Experiments}

\subsection{Experimental Setup}

\textbf{Models and Training Data.} We employ a diverse set of models, including Qwen2.5-0.5B-Instruct~\cite{qwen2025qwen25technicalreport} and Qwen2.5-Math-7B~\cite{yang2024qwen2}. For the 7B model, following Dr.GRPO~\cite{liu2025understanding}, we use MATH~\citep{hendrycks2021measuringmathematicalproblemsolving} Levels 3–5 (8,523 problems) as the training dataset, with a context length of 4K. Rewards are verifiable: ``1" for correct responses and ``0" for incorrect ones. We conduct ablation and analysis experiments primarily on the 0.5B model, which is trained on the GSM8K~\cite{cobbe2021trainingverifierssolvemath} dataset with a context length of 1K.

\textbf{Baselines and Implementation.} We primarily compare our method with GRPO~\cite{shao2024deepseekmathpushinglimitsmathematical}, TreeRL~\cite{hou2025treerl} and AttnRL~\cite{liu2025attention}. For our method (DEEP-GRPO), we set the loss weight $\lambda=1$, the branching count $M=8$, and $\gamma=2$ by default. Sensitivity analysis for the depth-bias coefficient $\gamma$ is provided in Appendix~\ref{app:gamma_sensitivity}. Further implementation details and hyperparameters are provided in Appendix \ref{app:implementation_details}. 

\textbf{Benchmarks.} We evaluate on GSM8K~\cite{cobbe2021trainingverifierssolvemath} and five challenging benchmarks following Dr.GRPO~\cite{liu2025understanding}: AIME24 (30 high-school olympiad problems)~\cite{li2024numinamath}, AMC (83 intermediate multiple-choice problems)~\cite{li2024numinamath}, MATH500 (500 problems from MATH)~\cite{lightman2023let}, Minerva~\cite{DBLP:conf/nips/LewkowyczADDMRS22} (272 graduate-level problems), and OlympiadBench (Oly.)~\cite{he2024olympiadbench} (675 high-difficulty problems). These benchmarks cover a broad spectrum of problem types and difficulties. We report Pass@1, setting the temperature to 0.0 and generating one answer per question.

\subsection{Main Results}

\begin{figure*}[ht]
    \centering
    \begin{subfigure}[b]{0.32\textwidth}
        \centering
        \includegraphics[width=\textwidth]{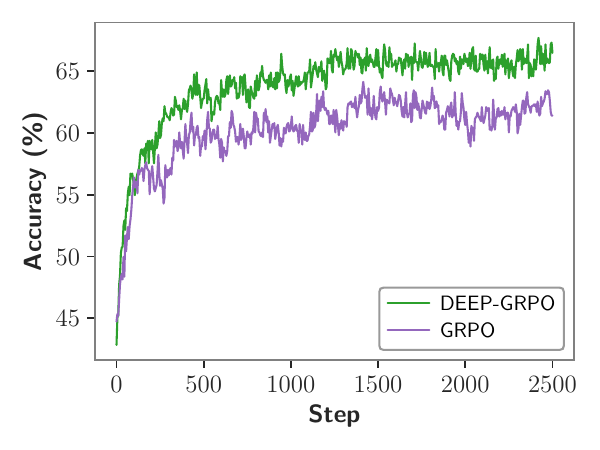}
        \caption{Test Accuracy}
        \label{fig:accuracy}
    \end{subfigure}
    \hfill
        \begin{subfigure}[b]{0.32\textwidth}
        \centering
        \includegraphics[width=\textwidth]{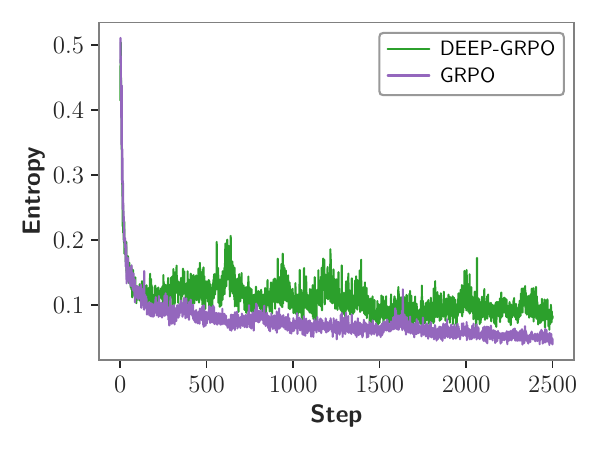}
        \caption{Entropy}
        \label{fig:entropy}
    \end{subfigure}
    \hfill
    \begin{subfigure}[b]{0.32\textwidth}
        \centering
        \includegraphics[width=\textwidth]{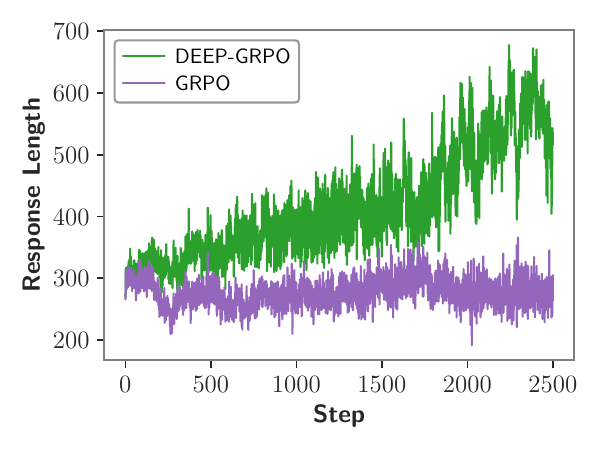}
        \caption{Response Length}
        \label{fig:length}
    \end{subfigure}
    \caption{Training dynamics on GSM8K. DEEP-GRPO achieves significant improvements in test accuracy compared to GRPO. Notably, our method maintains higher entropy and produces longer responses throughout the training.}
    \label{fig:training_dynamics}
\end{figure*}

We first compare our method against GRPO with varying group sizes ($n$) and tree-based baselines (TreeRL, AttnRL) on the GSM8K dataset (Table~\ref{tab:gsm8k_results}). While increasing $n$ from 8 to 64 for GRPO yields consistent gains (improving from 64.1\% to 66.2\%), marginal returns diminish significantly as $n$ increases from 16 to 64. Meanwhile, TreeRL achieves 65.5\%, outperforming GRPO ($n=8$) but falling short of GRPO with larger group sizes ($n \ge 16$). AttnRL performs better, achieving 67.0\%, likely because attention scores identify better branching points compared to entropy used by TreeRL. However, AttnRL still lags behind DEEP-GRPO. We hypothesize that branching based on entropy or attention scores may result in many branching points located at shallow positions, which are likely naturally covered by global root-level sampling, thereby diminishing the benefit of explicit branching. In contrast, DEEP-GRPO targets deep states that complement root sampling and achieves the best accuracy (67.7\%), suggesting that deep dense exploration provides a more effective pathway to performance improvement. As shown in Figure~\ref{fig:training_dynamics}, DEEP-GRPO maintains consistently higher policy entropy and produces longer responses compared to GRPO throughout the training. We hypothesize that this sustained exploration vitality prevents premature convergence and enables the model to refine its reasoning, contributing to the superior test accuracy. Further discussion on these dynamics is provided in Appendix~\ref{app:training_dynamics}. We also observe that tree-based methods suffer from training instability. We provide an analysis of this phenomenon in Appendix~\ref{app:tree_instability}.

\begin{table}[ht]
    \centering
    \caption{Performance comparison on GSM8K using Qwen2.5-0.5B-Instruct. We compare our method against GRPO (with varying group sizes $n$), TreeRL~\cite{hou2025treerl} and AttnRL~\cite{liu2025attention}.}
    \label{tab:gsm8k_results}
    \small
    \begin{tabular}{lc}
        \toprule
        \textbf{Method} & \textbf{Acc (\%)} \\
        \midrule
        GRPO ($n=8$)  & 64.1 \\
        GRPO ($n=16$) & 65.9 \\
        GRPO ($n=32$) & 66.0 \\
        GRPO ($n=64$) & 66.2 \\
        TreeRL~\cite{hou2025treerl} & 65.5 \\
        AttnRL~\cite{liu2025attention} & 67.0 \\
        \midrule
        \textbf{DEEP-GRPO (Ours)} & \textbf{67.7} \\
        \bottomrule
    \end{tabular}
\end{table}

Table \ref{tab:main_results} compares our method with recent strong baselines on five mathematical reasoning benchmarks (AIME24, AMC, MATH500, Minerva, Oly.). DEEP-GRPO achieves the highest average accuracy, reaching 54.0 and outperforming Dr.GRPO (51.4) and AttnRL (52.8). These results indicate that explicitly targeting deep states in failed trajectories during training provides effective complementary learning signals.

\begin{table*}[ht]
    \centering
    \caption{Main Results on Mathematical Benchmarks. We report the accuracy (\%) on AIME24, AMC, MATH500, Minerva, and Olympiad (Oly.) benchmarks.}
    \label{tab:main_results}
    \small
    \begin{tabular}{l|ccccc|c}
        \toprule
        \textbf{Model} & \textbf{AIME24} & \textbf{AMC} & \textbf{MATH500} & \textbf{Minerva} & \textbf{Oly.} & \textbf{Avg.} \\
        % \midrule
        % % \multicolumn{7}{l}{\textit{Size $\sim$ 1.5B}} \\
        % Qwen2.5-Math-1.5B~\cite{yang2024qwen2} & 16.7 & 43.4 & 61.8 & 15.1 & 28.4 & 33.1 \\
        % Qwen2.5-Math-1.5B-Instruct~\cite{yang2024qwen2} & 10.0 & 48.2 & 74.2 & 26.5 & 40.2 & 39.8 \\
        % Oat-Zero-1.5B (Dr. GRPO)~\cite{liu2025understanding} & 20.0 & 53.0 & 74.2 & 25.7 & 37.6 & 42.1 \\
        % AttnRL~\cite{liu2025attention} & 30.0 & 43.4 & 75.6 & 26.8 & 37.3 & 42.6 \\
        % \textbf{DEEP-GRPO-1.5B (Ours)} & 26.7 & 50.6 & 75.2 & 27.2 & 36.7 & \textbf{43.3} \\
        % % \textbf{DEEP-GRPO-1.5B* (Ours)} & 30.0 & 51.8 & 77.8 & 27.9 & 36.1 & \textbf{44.7} \\
        \midrule
        % \multicolumn{7}{l}{\textit{Size $\sim$ 7B}} \\
        Qwen2.5-Math-7B~\cite{yang2024qwen2} & 16.7 & 38.6 & 50.6 & 9.9 & 16.6 & 26.5 \\
        SimpleRL-Zero-7B~\cite{zeng2025simplerl} & 26.7 & 60.2 & 78.2 & 27.6 & 40.3 & 46.6 \\
        PRIME-Zero-7B~\cite{cui2025process} & 16.7 & 62.7 & 83.8 & 36.0 & 40.9 & 48.0 \\
        OpenReasoner-Zero-7B @ 3k~\cite{hu2025open} & 13.3 & 47.0 & 79.2 & 31.6 & 44.0 & 43.0 \\
        OpenReasoner-Zero-7B @ 8k~\cite{hu2025open} & 13.3 & 54.2 & 82.4 & 31.6 & 47.9 & 45.9 \\
        % Eurus-7B~\cite{yuan2024advancing} & 16.7 & 62.7 & 83.8 & 36.0 & 40.9 & 48.0 \\
        GPG-7B~\cite{chu2025gpg} & 33.3 & 65.0 & 80.0 & 34.2 & 42.4 & 51.0 \\
        Oat-Zero-7B (Dr. GRPO)~\cite{liu2025understanding} & 43.3 & 62.7 & 80.0 & 30.1 & 41.0 & 51.4 \\
        AttnRL~\cite{liu2025attention} & 40.0 & 63.9 & 81.4 & 34.6 & 44.3 & 52.8 \\
        \textbf{DEEP-GRPO-7B (Ours)} & 46.7 & 65.1 & 81.6 & 33.8 & 42.6 & \textbf{54.0} \\
        \bottomrule
    \end{tabular}
\end{table*}

\subsection{Generalization to Agent Scenario}

To further evaluate the generalizability of DEEP-GRPO beyond mathematical reasoning, we conduct experiments on Multi-Hop QA, where the model acts as an agent that iteratively invokes a retriever to gather evidence across multiple hops before producing the final answer. Following the experimental setup of Tree-GRPO~\cite{ji2025tree}, we use Qwen2.5-3B as the base model, adopt the same training set and reward function, and train for one epoch. 

\textbf{Results.} As shown in Table~\ref{tab:agent_results}, DEEP-GRPO achieves the best performance on both HotpotQA~\cite{yang2018hotpotqa} and 2WikiMultiHopQA~\cite{ho2020constructing}. Notably, Tree-GRPO, which performs random branching from intermediate states, outperforms root-only methods such as GRPO~\cite{shao2024deepseekmathpushinglimitsmathematical} and GSPO~\cite{zheng2025group}, validating the effectiveness of intermediate-state exploration in agent scenarios. DEEP-GRPO further improves upon Tree-GRPO, suggesting that utility-guided pivot selection and dual-stream optimization provide more effective exploration and policy learning.

\begin{table}[ht]
    \centering
    \caption{Performance comparison on multi-hop QA agent benchmarks. We compare DEEP-GRPO with GRPO, GSPO, and Tree-GRPO on HotpotQA and 2WikiMultiHopQA using Qwen2.5-3B as the base model.}
    \label{tab:agent_results}
    \small
    \begin{tabular}{lcc}
        \toprule
        \textbf{Method} & \textbf{HotpotQA} & \textbf{2WikiMultiHopQA} \\
        \midrule
        GRPO & 39.0 & 36.3 \\
        GSPO & 40.2 & 39.8 \\
        Tree-GRPO & 42.4 & 43.7 \\
        \textbf{DEEP-GRPO} & \textbf{45.1} & \textbf{43.9} \\
        \bottomrule
    \end{tabular}
\end{table}

\subsection{Comparison of Different Auxiliary Loss Weights}

We investigate the impact of the weighting coefficient $\lambda$ in Eq.~\ref{eq:optimization_objective}, which balances the main chain loss and the auxiliary chain loss. We vary $\lambda \in \{0.1, 0.5, 1.0, 1.5, 2.0, 10.0\}$ on GSM8K.

\textbf{Results.} As shown in Table~\ref{tab:different_loss_weight}, $\lambda=1.0$ achieves the best performance. When $\lambda$ is too small, such as $\lambda=0.1$, the auxiliary chain signal is overwhelmed by the main chain signal, reducing the method closer to standard GRPO. Conversely, when $\lambda$ is too large, such as $\lambda=10.0$, the auxiliary objective dominates optimization and hinders global policy learning. These results suggest that a balanced weighting between global policy learning and local corrective updates is important for DEEP-GRPO.

\begin{table}[ht]
    \centering
    \caption{Performance comparison with different auxiliary loss weights $\lambda$.}
    \label{tab:different_loss_weight}
    \small
    \renewcommand{\arraystretch}{1.0}
    \begin{tabular}{c|cccccc}
        \toprule
        $\lambda$ & $0.1$ & $0.5$ & $1.0$ & $1.5$ & $2.0$ & $10.0$ \\
        \midrule
        Acc. & 65.4 & 66.6 & \textbf{67.7} & 67.1 & 67.6 & 65.9 \\
        \bottomrule
    \end{tabular}
\end{table}

\subsection{Comparison of Different Exploration Budget Allocation Strategies}

We further study how the allocation of the auxiliary exploration budget affects DEEP-GRPO. We compare three configurations with varying branching widths and pivot counts: single-pivot with 4 branches ($P_1B_4$) and 8 branches ($P_1B_8$), as well as new double-pivot case ($P_2B_4$), where 2 pivots are selected per unsuccessful trajectory, each with 4 branches.

\textbf{Results.} As shown in Figure \ref{fig:scalability_of_exploration_budget}, under the same training time, $P_1B_8$ achieves the best performance, followed by $P_2B_4$ and then $P_1B_4$. This offers two key insights:

\begin{figure}[ht]
    \centering
    \includegraphics[width=0.842\linewidth]{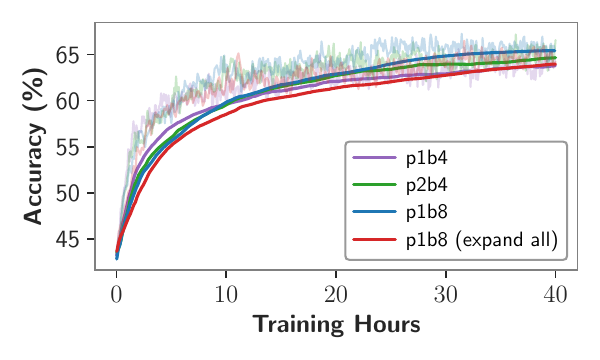}
    \caption{Comparison of auxiliary budget allocation strategies across pivot counts and branching widths.}
    \label{fig:scalability_of_exploration_budget}
\end{figure}

First, under the same wall-clock training time, the performance gain from $P_1B_4$ to $P_1B_8$ highlights the importance of a sufficient branching width. This suggests that for ``hard" states, a smaller budget (4 rollouts) is often insufficient to discover a correct solution suffix. Doubling the local budget to 8 significantly increases the discovery probability.

Second, the comparison between $P_1B_8$ and $P_2B_4$ provides a crucial insight into budget allocation. Notably, both configurations use the same total exploration budget (8 rollouts). The superior performance of $P_1B_8$ indicates that our ``dense exploration" strategy, which concentrates the budget on a single, critical pivot, is more effective than distributing it sparsely across multiple pivots, the approach adopted by tree-based strategies. This advantage stems from two key factors. First, by concentrating the rollouts, this strategy significantly increases the probability of discovering a correct solution suffix for difficult states. Second, these concentrated rollouts yield a more stable local baseline for the policy update, leading to more effective learning.

Furthermore, we conducted an ablation study by comparing $P_1B_8$ strategy with $P_1B_8$\textbf{-expand-all}, a variant that also branches from correct trajectories. The results show that $P_1B_8$ outperforms $P_1B_8$\textbf{-expand-all} under the same training time. This suggests that focusing the exploration budget on error-prone states is a more effective and compute-efficient strategy.

\subsection{Pass@K Analysis}

Since DEEP-GRPO is designed to improve exploration, we further evaluate whether it mitigates the ``limit-of-RLVR'' phenomenon~\cite{chen2026does}, which suggests that RLVR improves Pass@1 by biasing the model distribution toward rewarded reasoning paths but may narrow the reasoning capability boundary. Specifically, we evaluate the base model, the GRPO-trained model, and the DEEP-GRPO-trained model from the GSM8K experiments in Table~\ref{tab:gsm8k_results}, and report Pass@K by sampling $K$ responses for each test problem.

\textbf{Results.} As shown in Table~\ref{tab:pass_k}, GRPO achieves strong improvements at small $K$ values, but its advantage gradually diminishes as $K$ increases.  Notably, at $K=64$, GRPO is surpassed by the base model (86.8 for GRPO vs. 87.7 for the base model), reproducing the ``limit-of-RLVR'' phenomenon. In contrast, DEEP-GRPO consistently outperforms both GRPO and the base model across all $K$ values, including at $K=128$. This suggests that DEEP-GRPO mitigates the coverage narrowing issue to a notable extent compared with GRPO.

\begin{table}[ht]
\centering
\small
\renewcommand{\arraystretch}{1.0}
\begin{tabular}{c|ccc}  
    \toprule
    $K$ & Base Model & GRPO & DEEP-GRPO \\
    \midrule
    8   & 63.8 & 78.7 & \textbf{82.2} \\
    16  & 74.0 & 81.9 & \textbf{85.8} \\
    32  & 82.0 & 84.6 & \textbf{88.9} \\
    64  & 87.7 & 86.8 & \textbf{91.2} \\
    128 & 91.4 & 88.4 & \textbf{92.9} \\
    \bottomrule
\end{tabular}
\caption{Pass@K comparison of the base model, GRPO, and DEEP-GRPO.}
\label{tab:pass_k}
\end{table}

\section{Conclusion}

This work proposes DEEP-GRPO, a method designed to enhance the exploration capability of RL for LLMs. Unlike GRPO, which wastes compute on mastered paths, and tree-based methods that inefficiently disperse budget, our approach strategically concentrates exploration on pivots that complement the reach of root sampling. Specifically, we target deep states within unsuccessful trajectories that retain the potential for recovery, enabling the discovery of high-quality contrastive signals. By performing local dense resampling at these critical states, we compute stable local advantages and implement a dual-stream optimization that effectively resolves the weight instability issue arising from the dynamic sample sizes of the two streams. Our experiments demonstrate that DEEP-GRPO significantly outperforms baselines, establishing it as a highly effective method for LLM reinforcement learning.

\section*{Limitations}

Our evaluation mainly focuses on mathematical reasoning and multi-hop QA agent tasks, where outcome correctness can be automatically verified. Further work is needed to examine whether DEEP-GRPO generalizes to more open-ended tasks where such verifiable feedback is less readily available.

In addition, our empirical study is conducted on a limited set of model families and scales. Future work could evaluate DEEP-GRPO on larger models and more diverse model architectures.


\bibliography{custom}

\appendix

\section{Motivating Example}
\label{app:motivating_example}

\begin{figure*}[t]
    \centering
    \includegraphics[width=\textwidth]{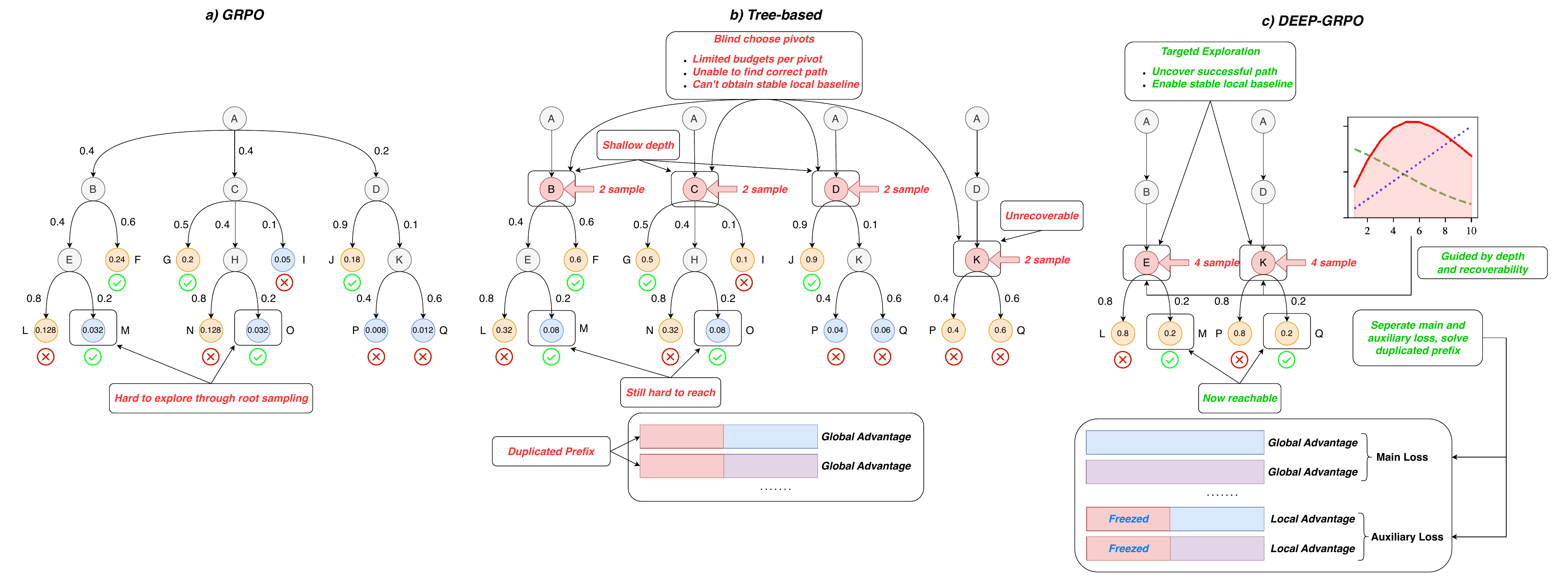}
    \caption{Motivating Example.}
    \label{fig:motivating_example}
\end{figure*}

To illustrate the limitations of existing exploration strategies and the necessity of our approach, we present a comparative analysis in Figure~\ref{fig:motivating_example}. We construct a probabilistic decision tree that represents a simplified response space for a given prompt. In this structure, each node corresponds to an intermediate state of the generation process, and the objective is to discover as many correct leaf nodes as possible (specifically, nodes $F$, $G$, $J$, $M$, and $O$, marked with green checks).

\paragraph{The Inefficiency of Root-Based Sampling (GRPO).}
As illustrated in Figure~\ref{fig:motivating_example}(a), GRPO initiate all rollouts from the root node $A$. A critical inefficiency arises because the distribution of the exploration budget is determined by the current policy. Consequently, the majority of the rollout budget naturally flows into high-probability paths (highlighted in orange, e.g., reaching nodes $F$, $G$, $J$, $L$, $N$).

This creates a ``rich-get-richer'' phenomenon where computational resources are monopolized by high-probability paths (e.g., $P(F)=0.24$, $P(G)=0.2$). In stark contrast, deep, low-probability states are statistically starved; for instance, reaching the deep state $M$ requires traversing a specific chain with a marginal probability of merely $0.4 \times 0.4 \times 0.2 = 0.032$. Consequently, simply increasing the root budget is inefficient, as additional samples disproportionately accumulate in the already saturated high-probability branches rather than penetrating the elusive, low-probability regions.

\paragraph{The Pitfalls of Tree-Based Exploration.}
Tree-based methods (Figure~\ref{fig:motivating_example}(b)) attempt to mitigate coverage issues by initiating rollouts from intermediate states. However, these methods typically adopt an unfocused strategy, dispersing the total computational budget across numerous pivots rather than concentrating it. This results in a diluted exploration budget for each branch (e.g., only 2 samples per pivot).

While pivoting brings the exploration deeper into the tree, the probability of reaching specific deep states often remains insufficient given the sparse samples. Taking state $O$ as an example: pivoting at node $C$ increases the access probability from the root-based marginal probability of $\mathbf{0.032}$ to $\mathbf{0.08}$. Despite this improvement, with a meager budget of 2 samples, the likelihood of successfully uncover state $O$ remains statistically minimal.

Furthermore, this approach introduces two critical optimization issues. First, the limited sample size (e.g., $N=2$) fails to provide a stable local baseline, introducing high variance into the advantage estimation. Second, these methods suffer from the duplicated prefix problem. Since multiple branches stem from the same trajectory (e.g., sharing the prefix $A \to C$), simply aggregating their losses leads to redundant reinforcement of the prefix.

\paragraph{Targeted Exploration with DEEP-GRPO.}
Our approach (Figure~\ref{fig:motivating_example}(c)) addresses these inefficiencies by utilizing a utility score to identify deep, error-prone, yet recoverable states as pivots (e.g., nodes $E$ and $K$). From these selected pivots, we conduct dense exploration. By targeting these specific deep regions, our method provides a critical complement to root sampling, uncovering training signals that are statistically inaccessible to root-based rollouts.

Take node $O$ as an example: by identifying its parent node as a high-utility pivot, we allocate a denser budget (e.g., 4 samples), raising the discovery probability significantly. This enables node $O$ to be uncovered to form a valid contrastive pair with its sibling $N$, providing the model with precise feedback on error-prone boundaries.

From an optimization perspective, this concentrated sampling allows us to compute a stable local baseline. Furthermore, we explicitly separate the learning objective into a main chain loss and an auxiliary chain Loss. When optimizing the auxiliary branches, we freeze the shared prefix, ensuring that gradient updates are driven solely by the quality of suffix. This effectively resolves the duplicated prefix problem.

\section{Pseudocode of DEEP-GRPO}
\label{app:pseudocode}

We present the full training procedure of DEEP-GRPO in
Algorithm~\ref{alg:tb_grpo}. The procedure consists of four stages:
main-chain sampling, targeted branching, policy optimization, and recoverability estimator updating.

\begin{algorithm}
    \SetAlgoLined
    \DontPrintSemicolon
    \caption{DEEP-GRPO Training Procedure}
    \label{alg:tb_grpo}

    \KwIn{Dataset $\mathcal{D}$, Policy $\pi_\theta$, Success Estimator $P_\phi$, Hyperparameters $\gamma, \lambda, K, G$}
    \KwOut{Optimized Policy $\pi_\theta$}

    \textbf{Initialize} replay buffer $\mathcal{M}$ for success estimator\;

    \While{not converged}{
        Sample a batch of queries $\mathbf{x} \sim \mathcal{D}$\;

        \tcp{1. Main Chain Sampling}
        Generate group $\mathcal{T}_{\text{main}} = \{\tau^1, \dots, \tau^G\}$ via $\pi_\theta(\cdot|\mathbf{x})$\;
        
        Compute rewards $R(\tau^i)$ for all $\tau^i \in \mathcal{T}_{\text{main}}$\;
        
        Compute global advantages $A^i_{\text{global}}$ using group statistics of $\mathcal{T}_{\text{main}}$\;

        Identify failed trajectories $\mathcal{T}_{\text{fail}} = \{ \tau^i \in \mathcal{T}_{\text{main}} \mid R(\tau^i) = 0 \} $\;

        \tcp{2. Targeted Branching}
        Initialize auxiliary set $\mathcal{T}_{\text{aux}} \gets \emptyset$\;

        \For{each failed trajectory $\tau^i \in \mathcal{T}_{\text{fail}}$}{
            Segment $\tau^i$ into $T$ candidate points\;
            
            Compute sampling prob $\mathcal{Q}(t)$ via Eq. \ref{eq:proposal_dist}\;
            
            Sample pivot step $t^*_i \sim \mathcal{Q}$\;

            \tcp{Local Resampling}
            Given prefix $s_{<t^*_i}$, generate $K$ auxiliary branches $\mathcal{T}_{\text{aux}}^{(i)} = \{\hat{\tau}^{i,1}, \dots, \hat{\tau}^{i,K}\}$\;

            Compute rewards $R(\hat{\tau}^{i,k})$ for $k=1 \dots K$\;

            \tcp{Update Buffer}
            $y \gets \mathbb{I}(\exists k, R(\hat{\tau}^{i,k}) = 1)$\;
            
            Store $(t^*_i/T, y)$ in $\mathcal{M}$\;

            Compute local advantages $A^{i,k}_{\text{local}}$ relative to $\mathcal{T}_{\text{aux}}^{(i)}$\;
            
            Add $\mathcal{T}_{\text{aux}}^{(i)}$ to $\mathcal{T}_{\text{aux}}$\;
        }

        \tcp{3. Optimization}
        Compute $\mathcal{L}_{\text{main}}$ and $\mathcal{L}_{\text{aux}}$ (with masking)\;
        
        Update $\theta \gets \theta - \eta \nabla_\theta (\mathcal{L}_{\text{main}} + \lambda \mathcal{L}_{\text{aux}})$\;

        \tcp{4. Estimator Update}
        Periodically update $P_\phi$ minimizing binary cross-entropy on $\mathcal{M}$\;
    }
\end{algorithm}

\section{Additional Details on Utility-Guided Pivot Sampling}
\label{app:pivot_sampling_details}

\subsection{Representative Sampling Configurations}
\label{app:pivot_sampling_configs}

\begin{figure*}[ht]
    \includegraphics[width=\textwidth]{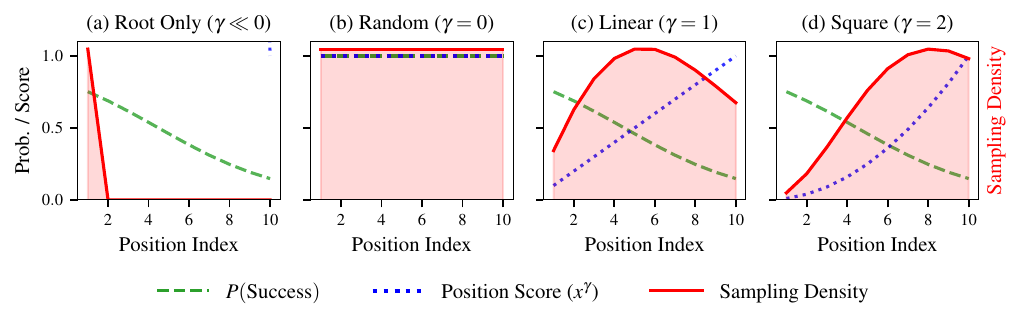}
    \caption{\textbf{Visualization of Pivot Sampling Distributions.} 
    We visualize the pivot sampling probability $Q(t)$ under representative configurations. 
    \textbf{Root Only} corresponds to $R(s_{<t}) \approx \mathrm{const}$ and a branching-value term $W(t)$ that strongly favors near-root positions, mimicking GRPO-style root sampling.
    \textbf{Random} corresponds to $R(s_{<t}) \approx \mathrm{const}$ and $W(t)=1$ (i.e., $\gamma=0$), yielding uniform sampling over candidate branching points. 
    \textbf{Linear} and \textbf{Square} correspond to DEEP-GRPO variants with $R(s_{<t})=R_\phi(s_{<t})$ and $W(t)=r_t^\gamma$ for $\gamma=1$ and $\gamma=2$, respectively, progressively shifting probability mass toward later positions while accounting for recoverability.}
    \label{fig:sampling_dist}
\end{figure*}

Figure~\ref{fig:sampling_dist} illustrates representative instantiations of Eq.~\ref{eq:proposal_dist}. 

In the root-only configuration, $R(s_{<t})\approx \mathrm{const}$, while $W(t)$ sharply favors early positions. 
Consequently, $Q(t)$ concentrates near the root, mimicking GRPO-style root sampling where complete trajectories are repeatedly generated from the prompt.

The random configuration corresponds to $R(s_{<t})\approx \mathrm{const}$ and $W(t)=1$, yielding nearly uniform sampling over candidate branching points. 

The linear and square configurations correspond to DEEP-GRPO variants with $R(s_{<t})=R_\phi(s_{<t})$ and $W(t)=r_t^\gamma$, where $\gamma=1$ and $\gamma=2$, respectively.
Larger $\gamma$ shifts more probability mass toward later prefixes, while $R_\phi(s_{<t})$ downweights prefixes with low expected recovery potential.

\subsection{Depth--Recoverability Relationship}
\label{app:depth_recoverability}

In Sec.~\ref{subsec:pivot_selection}, we use relative depth as a lightweight signal for estimating the recoverability of a failed trajectory prefix. Here, we provide additional statistical evidence for this design choice.

Specifically, we collect 636 failed trajectories, partition them into candidate branching positions, and evaluate recoverability at each position by sampling 8 rollout continuations from the corresponding prefix. A position is labeled as recoverable if at least one of these continuations reaches the correct answer. This procedure yields 6,346 position--recoverability pairs.

We then examine the relationship between the relative position ratio $r_t=t/T$ and recoverability. The relative position ratio is significantly negatively correlated with recoverability: the Spearman correlation between position ratio and mean recoverability is $\rho=-0.79$ ($p=1.4\times10^{-3}$). We further fit a logistic regression model to predict binary recoverability from $r_t$, which yields a coefficient $\hat{\beta}=-2.60$ ($z=-23.97$, $p\approx5.4\times10^{-127}$). The resulting depth-only model achieves $\mathrm{AUC}=0.68$, indicating that relative position carries meaningful predictive signal.

These results suggest that relative depth carries meaningful signal about the expected recoverability of failed-trajectory prefixes. Importantly, we do not require relative depth to precisely determine whether each individual prefix is recoverable. Instead, relative depth serves as a lightweight feasibility bias in the stochastic pivot sampling distribution, reducing the chance of allocating auxiliary rollouts to prefixes with low expected recovery potential.

\subsection{Online Update of $R_\phi$}
\label{app:rphi_update}

To update the recoverability estimator online, we maintain an experience buffer $\mathcal{M}$ storing tuples $(r_t,y_t)$ during training. We define the label $y_t=1$ if at least one of the $K$ auxiliary branches sampled from $s_{<t}$ successfully reaches the correct answer, and $y_t=0$ otherwise. The parameters $\phi=\{w,b\}$ of $R_\phi$ are updated periodically by minimizing the binary cross-entropy loss on $\mathcal{M}$.

\section{Sensitivity to the Depth-Bias Coefficient $\gamma$}
\label{app:gamma_sensitivity}

We investigate how the depth-bias coefficient $\gamma$ in the pivot sampling distribution affects model performance using Qwen2.5-0.5B-Instruct on GSM8K. A larger $\gamma$ shifts the pivot sampling distribution toward deeper states in the trajectory. This introduces a trade-off: deeper states provide signals that are more complementary to root-level sampling, but the likelihood of obtaining any correct solution among the branched auxiliary chains decreases.

\textbf{Results.} As shown in Table~\ref{tab:different_sampling_strategies}, when $\gamma=-20$, the distribution collapses to near root-only sampling, losing the benefit of deep exploration and performing the worst. As $\gamma$ increases beyond 2, the diminishing success rate of auxiliary chains outweighs the benefit of complementary signals. At $\gamma=2$, DEEP-GRPO achieves the best performance.

\begin{table}[ht]
    \centering
    \caption{Performance comparison with different depth bias coefficients $\gamma$.}
    \label{tab:different_sampling_strategies}
    \small
    \renewcommand{\arraystretch}{1.0}
    \begin{tabular}{c|cccccc}
        \toprule
        $\gamma$ & $-20$ & $1$ & $2$ & $3$ & $4$ & $5$ \\
        \midrule
        Acc. & 65.5 & 66.4 & \textbf{67.7} & 66.5 & 66.5 & 65.8 \\
        \bottomrule
    \end{tabular}
\end{table}

\section{Implementation Details}
\label{app:implementation_details}

Our training framework is built upon veRL~\cite{sheng2024hybridflow}. We adopt a strictly on-policy update strategy. Specifically, trajectories sampled in each batch are used to update the policy exactly once, ensuring that the model is always trained on data generated by the current policy. 

For the objective function, we utilize a batch-level token mean loss. This method sums the losses of all tokens across the entire batch and normalizes by the total token count. This ensures that every token contributes equally to the gradient, aligning with the approach in DAPO~\cite{yu2025dapoopensourcellmreinforcement}. Furthermore, for both our DEEP-GRPO method and the GRPO baseline, we compute policy updates exclusively using trajectories with non-zero advantages, consistent with recent practices~\cite{yu2025dapoopensourcellmreinforcement, lin2025cppoacceleratingtraininggroup}.

Regarding specific hyperparameters, we set the training batch size to 64. This indicates that at each training step, 64 unique prompts are sampled from the dataset, and for each prompt, 8 trajectories are generated. The model is trained with a learning rate of $1 \times 10^{-6}$ and a KL penalty coefficient of $1 \times 10^{-4}$.

\section{Training Dynamics}
\label{app:training_dynamics}

As shown in Figure~\ref{fig:training_dynamics}, DEEP-GRPO consistently outperforms GRPO in test accuracy (left), demonstrating superior efficiency in policy optimization.

\begin{figure*}[ht]
    \centering
    \begin{subfigure}[t]{0.46\linewidth}
        \centering
        \includegraphics[width=\linewidth]{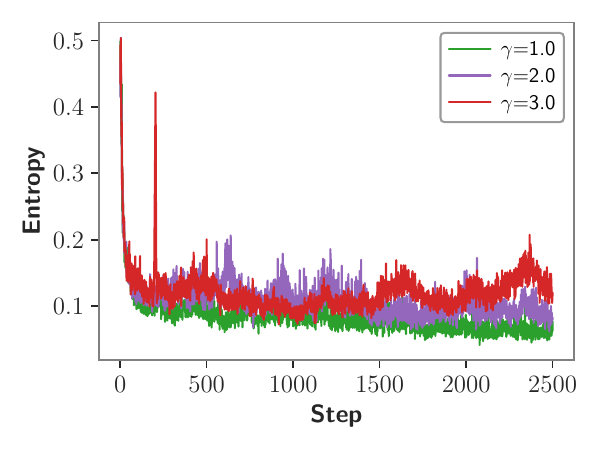}
        \caption{Policy Entropy}
        \label{fig:depth_entropy}
    \end{subfigure}
    \hfill
    \begin{subfigure}[t]{0.46\linewidth}
        \centering
        \includegraphics[width=\linewidth]{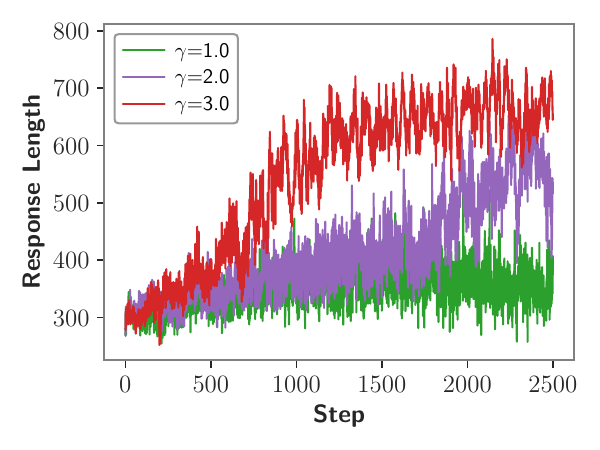} 
        \caption{Response Length}
        \label{fig:depth_length}
    \end{subfigure}
    
    \caption{Correlation between Depth Bias ($\gamma$) and Training Dynamics. Higher depth bias correlates with (a) \textbf{sustained policy entropy}, indicating maintained diversity, and (b) \textbf{increased response length}, suggesting the emergence of self-correction patterns.}
    \label{fig:depth_correlations}
\end{figure*}

We observed sustained higher policy entropy (middle). Unlike GRPO, which suffers from premature convergence, our method maintains greater exploration potential. To investigate the source of this sustained exploration, we analyzed the relationship between the depth of exploration and policy entropy (Figure~\ref{fig:depth_entropy}). We observe a positive correlation: as the depth bias increases, the policy entropy tends to remain higher.
This suggests that specifically targeting deep states for dense sampling effectively "injects" exploration vitality into the model.

Furthermore, we observe a notable increase in response length (right). As shown in Figure~\ref{fig:depth_length}, this increase is also positively correlated with the depth bias. We interpret this as the emergence of self-correction capabilities, as visually exemplified in Figure~\ref{fig:case_study}. By focusing exploration on deep, recoverable states, the model learns to ``double-check'' its logic, leading to longer chains that can leverage more test-time compute.

\begin{figure*}[ht]
    \centering
    \begin{tcolorbox}[
        enhanced,
        title={\textbf{Case Study: Self-Correction in DEEP-GRPO}},
        colframe=black!70,
        colback=white,
        coltitle=white,
        colbacktitle=black!70,
        attach boxed title to top left={xshift=5mm, yshift=-2mm},
        boxrule=0.5mm,
        sharp corners=south,
        arc=4mm,
        fonttitle=\bfseries\sffamily,
        fontupper=\small\sffamily
    ]
        \textbf{Problem:} Frankie watches TV... If he watched 7 hours of TV in all, how many 30-minute episodes did he watch on Wednesday?
        
        \rule{\linewidth}{0.5pt} 
        \vspace{0.2cm}
        
        \textbf{Model Response:}
        
        ... (omitted) ... On Thursday, he watched a 1-hour episode... and a 30-minute show...
        
        \begin{tcolorbox}[colback=myred, colframe=myred, arc=0mm, boxsep=0mm, left=2mm, right=2mm, top=1mm, bottom=1mm]
        \textbf{\textcolor{darkred}{[Initial Error Path]}} \\
        The total time: $2 + 30 + 1 + 120 = 162$ minutes. Since 1 hour = 60 minutes...
        \end{tcolorbox}
        
        \vspace{0.1cm}
        \begin{tcolorbox}[colback=myblue, colframe=myblue, arc=0mm, boxsep=0mm, left=2mm, right=2mm, top=1mm, bottom=1mm]
        \textbf{\textcolor{darkblue}{[Self-Correction Trigger]}} \\
        \textbf{However, let's re-evaluate:}
        \end{tcolorbox}
        \vspace{0.1cm}

        \begin{tcolorbox}[colback=mygreen, colframe=mygreen, arc=0mm, boxsep=0mm, left=2mm, right=2mm, top=1mm, bottom=1mm]
        \textbf{\textcolor{darkgreen}{[Corrected Reasoning]}} \\
        The correct solution:
        \begin{itemize}
            \item Monday and Tuesday: 1 hour each, 120 minutes.
            \item Wednesday: 30 minutes.
            \item Thursday: 1 hour and 30 minutes, 90 minutes.
            \item Friday: 2 episodes of 1 hour each, 120 minutes.
        \end{itemize}
        The total: $120 + 30 + 90 + 120 = 360$ minutes. Since 7 hours = 420 minutes...
        \end{tcolorbox}
        
        ... (calculation steps) ... 
        
        Therefore, the correct answer is \boxed{3}.
    \end{tcolorbox}
    \caption{The model initially falls into a unit mixing error (Red Zone). Crucially, it triggers a re-evaluation mechanism (Blue Zone) without external feedback, discarding the erroneous path and restarting the derivation with consistent units (Green Zone).}
    \label{fig:case_study}
\end{figure*}

% \begin{figure}[h]
%     \centering
%     \includegraphics[width=0.5\linewidth]{figures/depth_and_entropy.pdf}
%     \caption{Correlation between Depth Bias and Policy Entropy. Higher depth bias correlates with sustained entropy levels, indicating that deep-state exploration helps maintain policy diversity.}
%     \label{fig:depth_entropy_correlation}
% \end{figure}

\section{Instability of Tree-Based Methods}
\label{app:tree_instability}

\begin{figure*}[ht]
    \centering
    \begin{subfigure}[t]{0.46\textwidth}
        \centering
        \includegraphics[width=\linewidth]{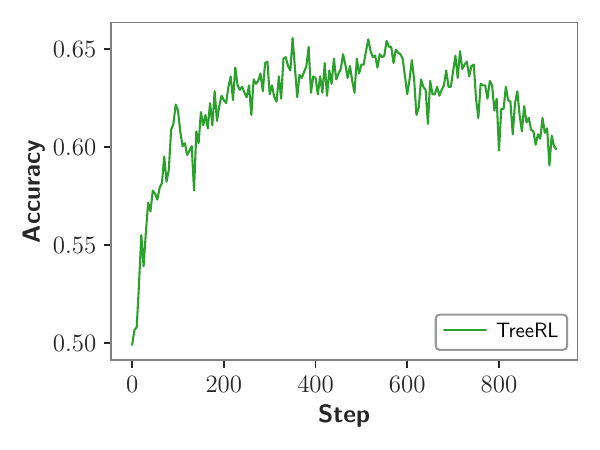} 
        \caption{Test Accuracy}
        \label{fig:instability_acc}
    \end{subfigure}
    \hfill
    \begin{subfigure}[t]{0.46\textwidth}
        \centering
        \includegraphics[width=\linewidth]{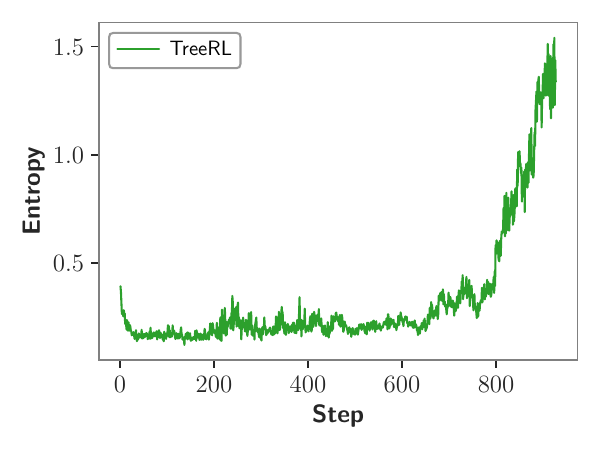}
        \caption{Policy Entropy}
        \label{fig:instability_entropy}
    \end{subfigure}

    \caption{Training instability in tree-based methods. (a) Test accuracy begins to degrade in later stages. (b) Policy entropy spikes sharply corresponding to the performance drop, indicating policy collapse.}
    \label{fig:tree_instability}
\end{figure*}

In our empirical evaluation, we observed a stability issue in the training dynamics of prior tree-based methods. Specifically, we noticed performance degradation in the later stages of training. We visualize this instability in Figure \ref{fig:tree_instability}, where the test accuracy begins to drop after reaching a peak, accompanied by a sharp spike in policy entropy.

We attribute this instability to two fundamental limitations inherent in methods that naively aggregate branched trajectories:

\paragraph{1. Optimization Bias from Prefix Duplication.}
Tree-based approaches typically treat branched trajectories as independent samples within a single optimization batch. However, since multiple branches ($\hat{\tau}_1, \dots, \hat{\tau}_K$) originate from the same intermediate pivot state, they share an identical prefix. Aggregating these trajectories without proper separation causes the gradients of the shared prefix to be accumulated $K$ times. This redundancy can destabilize the training, leading to the observed performance drop.

\paragraph{2. Biased Baseline Estimation.}
The second source of instability stems from the advantage estimation. Tree-based approaches typically compute a global baseline by aggregating all generated trajectories—both those sampled naturally from the root and those artificially branched from intermediate steps. We argue that this results in a biased baseline. Theoretically, the baseline should serve as an unbiased estimator of the value under the policy's natural distribution, $\mathbb{E}_{\tau \sim \pi(\cdot|\mathbf{x})}$. However, the set of branched trajectories represents an interventional distribution $\pi(\cdot|s_{\text{pivot}})$, which is artificially induced and skewed towards specific local states. Mixing these heterogeneous samples distorts the baseline, leading to inaccurate advantage estimates, thereby misguiding the optimization direction.

\section{Computational Efficiency Compared with Tree-based Methods}
\label{app:computational_efficiency}

Our method offers three budget advantages over tree-based methods (e.g., TreeRL~\cite{hou2025treerl}, TreeGRPO~\cite{ji2025tree}):
\begin{enumerate}
    \item \textbf{Adaptive Budget Allocation:} Unlike methods that construct branches for every main chain, we allocate the auxiliary budget exclusively to failed trajectories. As the policy improves, the number of failed trajectories decreases, naturally reducing the computational overhead of sampling auxiliary chains.
    \item \textbf{Concentrated Exploration:} Instead of scattering the sampling budget by branching at multiple positions (often selected randomly or via entropy), we select a single critical pivot per failed chain for dense refinement. This focused approach prevents the dilution of computational resources, allowing for intensive exploration of the state offering the highest exploration value.
    \item \textbf{Token Efficiency:} By explicitly targeting deep states (via our depth bias), our method effectively leverages the long existing prefix. Consequently, the model only needs to generate short suffixes to complete the trajectory. This incurs significantly lower token generation costs compared to branching at shallower positions.
\end{enumerate}

\section{More Related Work}
\label{app:more_related_work}

\subsection{Outcome-Based Reinforcement Learning for LLMs} 
% \lijie{Are the following methods all rely on root-based sampling? If yes, we can change the 2.1 section title to `Root-based sampling methods for LLM RL'}
Outcome-based methods, such as RLOO~\cite{ahmadian2024basicsrevisitingreinforcestyle}, GRPO~\cite{shao2024deepseekmathpushinglimitsmathematical}, GMPO~\cite{zhao2025geometric}, and GSPO~\cite{zheng2025group}, optimize policies using group-wise statistics without relying on a parametric value function. While efficient, these methods suffer from ``entropy collapse" as training progresses, leading to premature convergence~\cite{cui2025entropy}.
Recent works address this primarily by modifying the optimization objective.
For instance, DAPO~\cite{yu2025dapoopensourcellmreinforcement} proposes a Clip-Higher strategy that relaxes the upper bound of the probability ratio to facilitate the learning of exploration tokens. \citet{cui2025entropy} introduce Covariance Regularization to explicitly maintain policy entropy. \citet{song2025outcome} incorporate an exploration bonus into the optimization objective, such as a UCB-based term or a batch diversity penalty, to discourage redundant answers. Another line of work focuses on advantage shaping, where token advantages are modulated based on entropy~\cite{cheng2025reasoning} or perplexity and position~\cite{deng2025trial} to encourage exploration. Our work is orthogonal to these objective-level modifications. Rather than modifying the policy update method, we address the exploration problem from the data acquisition perspective, optimizing the sampling distribution to uncover high-value signals. Thus, our method can naturally complement these loss-based regularizers.

\end{document}